\documentclass[11pt]{article}
\PassOptionsToPackage{table}{xcolor}
\usepackage[preprint]{acl}
\usepackage{times}
\usepackage{latexsym}
\usepackage[T1]{fontenc}
\usepackage[utf8]{inputenc}
\usepackage{microtype}
\usepackage{inconsolata}
\usepackage{tgtermes}
\usepackage{fontspec}
\newfontfamily{\thaifont}{THSarabunNew.ttf}[Path=./]
\usepackage{graphicx}
\usepackage{booktabs}
\usepackage{amssymb}
\usepackage{amsmath}
\usepackage{tablefootnote}
\usepackage{siunitx}
\usepackage{adjustbox}
\usepackage{tikz}
\usepackage{pgfplots}
\pgfplotsset{compat=1.18}
\title{JaiTTS: A Thai Voice Cloning Model}

\author{
\textbf{Jullajak~Karnjanaekarin}\textsuperscript{1}\thanks{Equal contribution} \quad
\textbf{Pontakorn~Trakuekul}\textsuperscript{1}\footnotemark[1] \\
\textbf{Narongkorn~Panitsrisit}\textsuperscript{1} \quad
\textbf{Sumana~Sumanakul}\textsuperscript{1} \quad
\textbf{Vichayuth~Nitayasomboon}\textsuperscript{1} \\
\textbf{Nithid~Guntasin}\textsuperscript{3}\thanks{Work performed during internship at Jasmine Technology Solution} \quad
\textbf{Thanavin~Denkavin}\textsuperscript{3}\footnotemark[2] \quad
\textbf{Attapol~T.~Rutherford}\textsuperscript{1,2} \\[0.8ex]
\textsuperscript{1}Jasmine Technology Solution \\
\textsuperscript{2}Department of Linguistics, Chulalongkorn University \\
\textsuperscript{3}Sirindhorn International Institute of Technology \\[0.8ex]
\texttt{jts.ai.team@gmail.com}
}

\begin{document}

\maketitle
\vspace{-1.2em}

\begin{abstract}
We present JaiTTS-v1.0, a state-of-the-art Thai voice cloning text-to-speech model built through continual training on a large Thai-centric speech corpus. The model architecture is adapted from VoxCPM, a tokenizer-free autoregressive TTS model. JaiTTS-v1.0 directly processes numerals and Thai-English code-switching, which is very common in realistic settings, without explicit text normalization. We test the models on short-  and long-duration speech generation, which reflects many real-world use cases. 
 JaiTTS-v1.0 achieves a state-of-the-art CER of 1.94\%, surpassing the human ground truth of 1.98\% for short-duration tasks while performing on par with human ground truth for long-duration tasks. In human judgment evaluations, our model wins 283 of 400 pairwise comparisons against commercial flagships, with only 58 losses. Our code and demo are available at \url{https://github.com/JTS-AI-Team/JaiTTS}.
\end{abstract}

\begin{figure*}[!t]
\centering
\includegraphics[width=\textwidth]{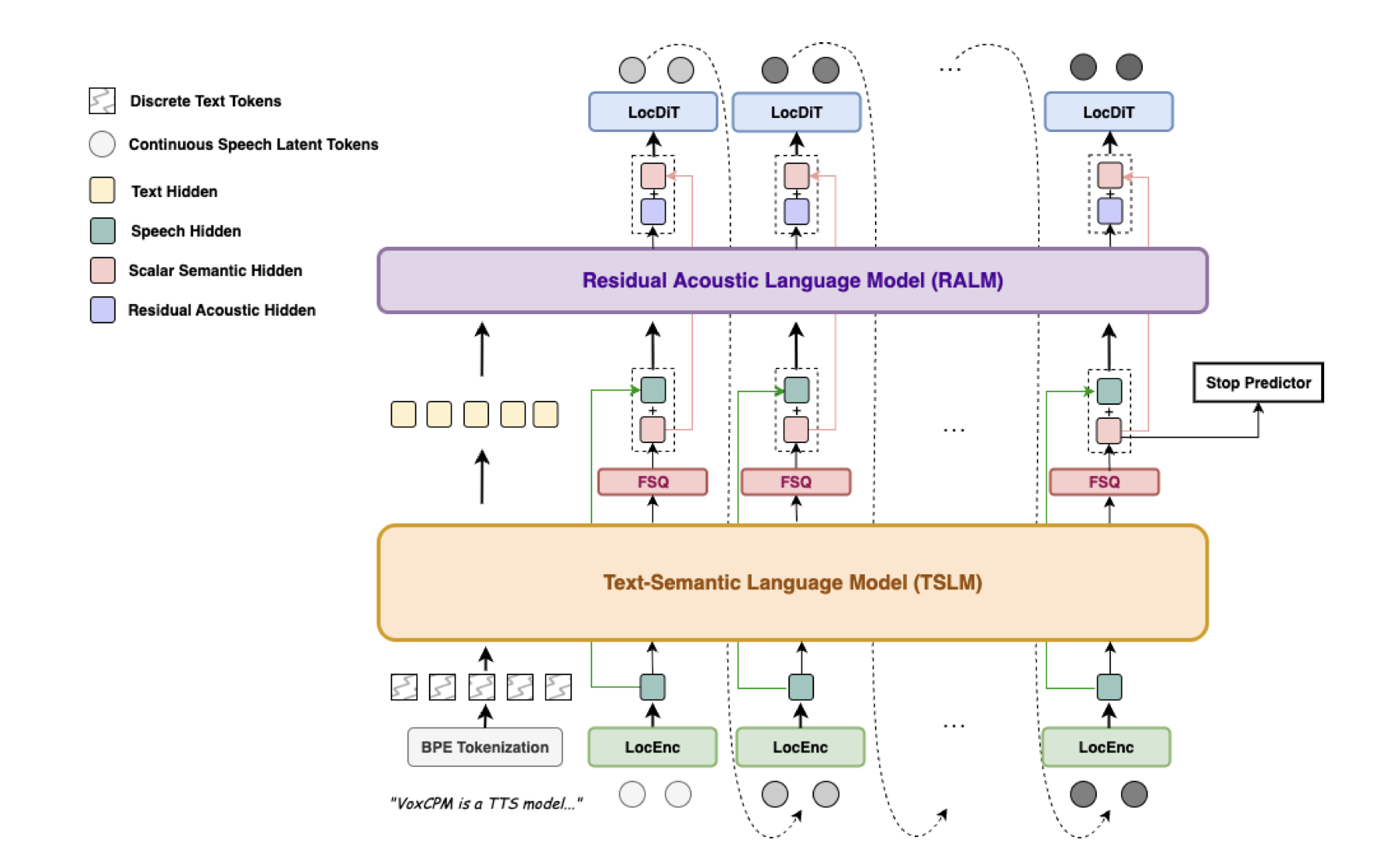}
\caption{Architecture of VoxCPM, the backbone of JaiTTS-v1.0. The Text-Semantic Language Model (TSLM) plans semantic-prosodic content from text and reference speech embeddings; a Finite Scalar Quantization (FSQ) layer compresses the TSLM hiddens into a scalar semantic skeleton; the Residual Acoustic Language Model (RALM) restores fine acoustic detail; and the Local Diffusion Transformer (LocDiT) decodes the result into continuous speech latent patches. A stop-prediction head consumes the post-FSQ hidden state to signal end-of-sequence. Figure adapted from \citet{voxcpm}.}
\label{fig:voxcpm_arch}
\end{figure*}

\section{Introduction}
Recent advancements in zero-shot text-to-speech (TTS) synthesis have revolutionized voice cloning capabilities by allowing highly natural audio generation from unseen reference voices. However, the majority of open source models and commercial systems are heavily optimized for English. Existing open source multilingual models, such as the Qwen3-TTS family \cite{qwen3tts}, provide strong foundational performance but occasional pronunciation and prosodic errors can still be observed possibly because Thai represents a small fraction of the training data.  Meanwhile, dedicated Thai-specific systems, such as ThonburianTTS \cite{thonburiantts}, remain limited in their zero-shot speaker cloning capabilities and struggle with extended speech generation because it was trained on the short-utterance-dominated GigaSpeech2 dataset \cite{gigaspeech2}. Furthermore, traditional Thai text-to-speech pipelines heavily rely on complex text normalization processes to handle numbers and the frequent Thai and English code-switching found in everyday communication.

Beyond these multilingual and Thai-specific systems, a recent line of autoregressive TTS work frames speech generation as next-token prediction over discrete neural audio-codec tokens. Notable examples include LLaSA \cite{llasa} and Inworld TTS-1 \cite{ttsone}: LLaSA introduces X-codec2 as its speech tokenizer, while TTS-1 builds a high-resolution codec on the X-codec2 architecture, which is aligned well with decoder-only training \cite{xcodec}.  However, the publicly reported X-codec2 training data are multilingual but do not appear to include Thai. Consequently, its ability to reconstruct and model Thai-specific phonetic contrasts, including lexical tone and Thai consonant clusters, remains empirically unverified.

To bridge this gap, we present a state-of-the-art Thai voice cloning system capable of natural and accurate speech synthesis. We train JaiTTS-v1.0 on a Thai-centric speech corpus, adapting it from VoxCPM \cite{voxcpm}, a tokenizer-free architecture that bypasses external speech codecs entirely by performing hierarchical semantic-acoustic modeling over continuous speech latents with semi-discrete residual representations. Our system directly processes raw text to seamlessly synthesize Thai and English code-switching and numeric inputs without requiring prior text transformation. This capability simplifies the deployment pipeline while maintaining pronunciation accuracy.

We introduce a novel benchmark categorizing target texts into short segments of 1 to 15 seconds and long segments of 16 to 30 seconds. Our results demonstrate state-of-the-art CER and competitive speaker similarity for JaiTTS-v1.0 across both short- and long-duration tasks. Furthermore, JaiTTS-v1.0 achieves a Real-Time Factor (RTF) of 0.1136, which corresponds to speech generation nearly 9$\times$ faster than real-time. In human judgment comparisons against strong commercial models, JaiTTS-v1.0 wins 283 of 400 pairwise comparisons, with 59 ties and 58 losses.

\section{Model Architecture}

JaiTTS-v1.0 is adapted from VoxCPM, a tokenizer-free autoregressive TTS model that predicts continuous speech latents directly, without relying on an external speech codec.

\paragraph{Overview.} Given an input text and a reference waveform, JaiTTS-v1.0 generates the target speech as a sequence of continuous latent patches produced by a separately trained causal audio variational autoencoder (VAE). Generation proceeds autoregressively, with each patch decoded by a hierarchical pipeline of four core components illustrated in Figure~\ref{fig:voxcpm_arch}. The Text-Semantic Language Model (TSLM) plans the semantic and prosodic content of the utterance from text and reference acoustic context. A Finite Scalar Quantization (FSQ) layer compresses the TSLM output into a semi-discrete skeleton that stabilizes the planning signal. The Residual Acoustic Language Model (RALM) operates on this skeleton to enhance speaker similarity and acoustic detail beyond what the discrete bottleneck can represent. Finally, the Local Diffusion Transformer (LocDiT) denoises Gaussian samples into the next continuous latent patch, conditioned on the semi-discrete skeleton from FSQ and the residual acoustic details from RALM.

\paragraph{Local Audio Encoder.} The LocEnc compresses the historical VAE latent patches $\mathbf{Z}_{<i}$ into a sequence of compact acoustic embeddings $\mathbf{E}_{<i}$. These embeddings provide both the TSLM and the RALM with an efficient acoustic context that supports speaker preservation.

\paragraph{Text-Semantic Language Model.} The TSLM is a decoder-only Transformer initialized from MiniCPM-4 \cite{minicpm4} that conditions on the BPE-tokenized text $\mathbf{T}$ and the historical acoustic context $\mathbf{E}_{<i}$ to produce a continuous semantic-prosodic representation $h_i^{\text{TSLM}}$. By inheriting the pre-trained language model's contextual understanding, the TSLM produces continuous representations that jointly capture the semantic content to be spoken and the prosodic structure of its delivery.

\paragraph{Finite Scalar Quantization.} The FSQ layer \cite{fsq} projects $h_i^{\text{TSLM}}$ onto a structured lattice through per-dimension scalar quantization, yielding a semi-discrete skeleton $h_i^{\text{FSQ}}$. For each dimension $j$,
\begin{equation}
h^{\text{FSQ}}_{i,j} = \Delta \cdot \mathrm{clip}\!\left(\mathrm{round}\!\left(\frac{h^{\text{TSLM}}_{i,j}}{\Delta}\right),\, -L,\, L\right),
\end{equation}
where $\Delta$ is the quantization step and $L$ bounds the discrete range. Differentiability is maintained through the straight-through estimator, allowing FSQ to act as a regularizer on the hidden state space rather than as a target vocabulary.

\paragraph{Residual Acoustic Language Model.} The RALM is a decoder-only Transformer that specializes in acoustic expressivity and speaker characteristics. It conditions on the text-side TSLM hidden states $H^{\text{TSLM}}_{\text{text}}$, the historical FSQ skeletons $H^{\text{FSQ}}_{<i}$, and the historical acoustic embeddings $\mathbf{E}_{<i}$ to produce a continuous residual representation
\begin{equation}
h^{\text{res}}_i = \mathrm{RALM}\!\left(H^{\text{TSLM}}_{\text{text}},\; H^{\text{FSQ}}_{<i} \oplus \mathbf{E}_{<i}\right),
\end{equation}
which complements the skeleton with speaker characteristics that the discrete bottleneck cannot represent. This explicit separation between semantic planning (TSLM) and acoustic refinement (RALM) avoids the task entanglement that arises in purely continuous models.

\paragraph{Local Diffusion Transformer.} The LocDiT is a bidirectional Transformer that decodes the next latent patch $z_i$ via a flow-matching denoising process. It receives the hierarchical conditioning $h_i^{\text{final}} = h_i^{\text{FSQ}} + h_i^{\text{res}}$, the previous patch $z_{i-1}$, and a diffusion timestep. Including the previous patch frames each local decoding as an outpainting task and improves cross-patch continuity.

\paragraph{Stop Predictor.} A lightweight head consumes the FSQ skeleton $h_i^{\text{FSQ}}$ and produces a binary logit indicating whether the current patch terminates the sequence.

\paragraph{Training Objectives.} The four modules and the FSQ layer are optimized jointly, end-to-end. Let $z_i^0$ denote the ground-truth latent for patch $i$, and define its noised version under the diffusion schedule $(\alpha_t, \sigma_t)$ for $t \in [0, 1]$ as
\begin{equation}
z_i^t = \alpha_t z_i^0 + \sigma_t \epsilon, \quad \epsilon \sim \mathcal{N}(0, I).
\end{equation}
The LocDiT velocity field $v_\theta$ is trained by conditional flow matching to regress the time derivative of this interpolation, conditioned on the combined signal $h_i^{\text{final}} = h_i^{\text{FSQ}} + h_i^{\text{res}}$ and the previous patch $z_{i-1}$:
\begin{multline}
\mathcal{L}_{\text{FM}} = \mathbb{E}_{t, z_i^0, \epsilon}\Big[\big\| v_\theta(z_i^t, t, h_i^{\text{final}}, z_{i-1}) \\
- \tfrac{d}{dt}(\alpha_t z_i^0 + \sigma_t \epsilon) \big\|^2\Big].
\end{multline}
The stop predictor $s_\theta$ is trained with a binary cross-entropy loss against the indicator $\mathbb{1}[\text{token } i \text{ is the last}]$:
\begin{multline}
\mathcal{L}_{\text{Stop}} = \mathbb{E}_{i \sim \text{sequence}}\Big[\text{BCE}\big(s_\theta(h_i^{\text{FSQ}}), \\
\mathbb{1}[\text{token } i \text{ is the last}]\big)\Big].
\end{multline}
The combined objective
\begin{equation}
\mathcal{L} = \mathcal{L}_{\text{FM}} + \lambda \mathcal{L}_{\text{Stop}}
\end{equation}
propagates through all four modules and the FSQ layer via the straight-through estimator, allowing the semantic-planning and acoustic-rendering pathways to be optimized together rather than as separately trained tokenizer and decoder stages. To enable classifier-free guidance at inference, the language-model conditioning fed into the LocDiT is randomly dropped with probability 0.1 during training.

\section{Experimental Setup}

\subsection{Datasets}
JaiTTS-v1.0 is trained on a Thai-centric speech corpus of approximately 10,000 hours. We develop an internal data pipeline to process, curate, and prepare the speech data for continual training. The corpus combines broad-domain general speech with content from four targeted verticals to provide both stylistic breadth and domain-specific terminology exposure. The general portion covers a wide range of conversational and formal styles drawn from sources such as podcasts, while the domain-specific portion spans \emph{Finance}, \emph{Healthcare}, \emph{Education} and \emph{Law}. Recording conditions are equally diverse: studio-grade recordings supply clean acoustic conditions, while crowd-sourced speech contributes natural prosodic variation and speaker diversity. All audio is paired with a transcript obtained through an automatic speech recognition pipeline followed by multi-step post-processing and verification to ensure transcription accuracy.

The evaluation set is split into two subsets based on the duration of the ground-truth audio: short-duration (1-15 seconds) and long-duration (16-30 seconds).
The curation pipeline for our short-duration evaluation follows the methodology established by Seed-TTS \cite{seedtts}. To ensure the highest audio quality for benchmarking, we first filter the entire Thai Common Voice \cite{commonvoice} test set using the DNSMOS Pro \cite{dnsmospro} metric, retaining only recordings with a score exceeding 3.9. We then randomly sample 1,000 utterances from this filtered pool. Finally, the audio files are denoised, and all leading and trailing silence is trimmed. This set serves to measure the model’s ability to capture speaker identity and produce high-quality audio in standard, concise segments.

To evaluate voice cloning stability over extended durations, we introduce a long-duration evaluation set. We draw 231 test cases from YouTube to represent more diverse and challenging acoustic environments than standard public corpora. We verify all transcriptions manually and correct any errors in the source text to ensure data quality. This benchmark evaluates the model’s ability to maintain prosodic consistency and avoid degradation over longer synthesis windows.

We apply a specific text processing strategy to test the model’s direct synthesis capabilities. We convert all Thai transliterated terms back to their original English spelling and transform Thai number words into Arabic numerals. This ensures that evaluation conditions mirror realistic input, where Thai-English code-switching and Arabic numerals appear in raw text.

\subsection{Baseline Models}
We evaluate JaiTTS-v1.0 against a human baseline, three open-source state-of-the-art systems, and one commercial Thai TTS system. The open-source baselines include Qwen3-TTS-0.6B, Qwen3-TTS-1.7B, and ThonburianTTS. The Qwen3-TTS baselines are autoregressive dual-track language-model TTS systems with dedicated speech tokenizers \cite{qwen3tts}, while ThonburianTTS is a Thai model based on F5-TTS \cite{thonburiantts}. We also evaluate Kaitom Voice V3\footnote{\url{https://iapp.co.th/th/blog/kaitom-voice-thai-text-to-speech-v3}}, a commercial Thai text-to-speech model developed by iApp Technology. Note that we test Kaitom Voice V3 during its Alpha phase, and its performance may improve in future releases.

\subsection{Evaluation Metrics}
We use Character Error Rate (CER) for intelligibility and stability, and Speaker Similarity (SIM) for acoustic correspondence on a scale of 0 to 1. We compute CER by transcribing the generated audio with the Typhoon-Whisper-Large-v3 model \cite{typhoonasr}, chosen for its strong Thai speech recognition accuracy and automatic text normalization capabilities. To align with the normalized form produced by the ASR, we do not compute CER against the raw input text fed to the TTS models. Instead, we apply matching normalization to the reference text: Arabic numerals are converted to Thai number words, English words are transliterated into Thai script, and the Thai repetition marker mai yamok ({\thaifont ๆ}) is expanded into the duplicated word (e.g., {\thaifont ต่างๆ} $\rightarrow$ {\thaifont ต่างต่าง}).

For SIM, we compute the cosine similarity of speaker embeddings extracted using a WavLM-Large model fine-tuned for speaker verification \cite{wavlm, wavlm-spk}. During the benchmarking of ThonburianTTS, we set the inference parameters \texttt{cfg\_strength} to 2.5 and \texttt{nfe\_step} to 32. For JaiTTS-v1.0, we set \texttt{cfg\_value} to 2.5 and \texttt{inference\_timesteps} to 10. We evaluate results from five independent runs to obtain the average scores. 

\section{Experimental Results}

\begin{table*}[!t]
\centering
\caption{Evaluation results on short- and long-duration Thai voice cloning benchmarks. Bold values indicate the best performance among synthesized models, while underlined values represent the second-best results.}
\label{tab:evaluation_results}
\begin{tabular}{lcccc}
\toprule
\textbf{Model} & \multicolumn{2}{c}{\textbf{Short (1 to 15s)}} & \multicolumn{2}{c}{\textbf{Long (16 to 30s)}} \\
\cmidrule(lr){2-3} \cmidrule(lr){4-5}
 & \textbf{CER(\%)} $\downarrow$ & \textbf{SIM} $\uparrow$ & \textbf{CER(\%)} $\downarrow$ & \textbf{SIM} $\uparrow$ \\
\midrule
Human (Ground Truth) & 1.98 & 0.61 & 2.47 & 0.83 \\
\midrule
Qwen3-TTS-0.6B & 3.14 & \textbf{0.62} & 6.10 & \textbf{0.79} \\
Qwen3-TTS-1.7B & 2.56 & \textbf{0.62} & \underline{3.64} & \underline{0.78} \\
ThonburianTTS & 6.26 & 0.48 & -- & -- \\
Kaitom Voice V3 (Alpha) & \underline{2.34} & \underline{0.59} & 5.81 & \textbf{0.79} \\
JaiTTS-v1.0 & \textbf{1.94} & \textbf{0.62} & \textbf{2.55} & 0.76 \\
\bottomrule
\end{tabular}
\end{table*}

In the short-duration benchmark, JaiTTS-v1.0 achieves a CER of 1.94\%, outperforming all evaluated baselines and marginally surpassing the human ground truth of 1.98\% (Table~\ref{tab:evaluation_results}), likely because synthesized audio tends to be cleaner than natural human speech. JaiTTS-v1.0 also maintains a competitive SIM of 0.62 Among the baselines, Kaitom Voice V3 (Alpha) performs best with a CER of 2.34\%, while the Thai-specific ThonburianTTS struggles in zero-shot scenarios, yielding a CER of 6.26\% and a SIM of 0.48.

In the long-duration benchmark, Qwen3-TTS-0.6B drops to a CER of 6.10\%, while the performance of Qwen3-TTS-1.7B falls to a CER of 3.64\%. Kaitom Voice V3 (Alpha) reaches a CER of 5.81\% and a SIM of 0.79. ThonburianTTS fails to generate sensible output, possibly because it is not trained on longer speech snippets; therefore, it is omitted from the analysis. Conversely, JaiTTS-v1.0 achieves a CER of 2.55\%, which is closely comparable to the human reference of 2.47\%, and outperforms the other strong baselines in our experiment setups.

\begin{table}[h]
\centering
\small
\caption{Real-Time Factor (RTF) comparison across models. Lower values indicate faster synthesis. Bold denotes the best result.}
\label{tab:rtf_results}
\begin{tabular}{lc}
\toprule
\textbf{Model} & \textbf{RTF} $\downarrow$ \\
\midrule
\multicolumn{2}{l}{\textit{Autoregressive (AR)}} \\
Qwen3-TTS-0.6B & 1.5092 \\
Qwen3-TTS-1.7B & 1.5409 \\
JaiTTS-v1.0 & \textbf{0.1136} \\
\midrule
\multicolumn{2}{l}{\textit{Non-Autoregressive (NAR)}} \\
ThonburianTTS & 0.1150 \\
\bottomrule
\end{tabular}
\end{table}

For computational efficiency, we compare Real-Time Factor (RTF) across systems under the same hardware and evaluation conditions (Table~\ref{tab:rtf_results}). The models vary widely in their computational efficiency. The Qwen3-TTS models score greater than 1.5 on RTFs, while ThonburianTTS reaches 0.1150. JaiTTS-v1.0 achieves an RTF of 0.1136, approximately 13$\times$ faster than the Qwen3-TTS models while remaining well within real-time constraints. JaiTTS-v1.0 is the fastest model compared to the other baselines. 

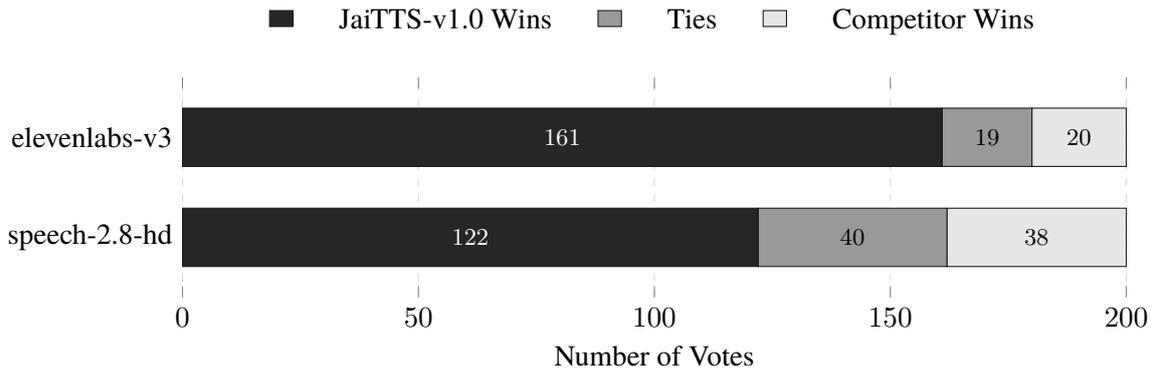
\begin{figure*}[!t]
\centering
\begin{tikzpicture}
\begin{axis}[
    xbar stacked,
    width=14cm,
    height=4.5cm,
    enlarge y limits={abs=0.6},
    xmin=0, xmax=200,
    ytick={1,2},
    yticklabels={speech-2.8-hd, elevenlabs-v3},
    ytick style={draw=none},
    xlabel={Number of Votes},
    legend style={
        at={(0.5,1.15)}, 
        anchor=south, 
        legend columns=-1, 
        draw=none, 
        font=\normalsize,
        column sep=15pt 
    },
    bar width=22pt,
    nodes near coords,
    nodes near coords align={center},
    axis line style={draw=none},
    tick label style={font=\normalsize},
    xtick={0, 50, 100, 150, 200},
    grid=major,
    grid style={dashed, gray!30}
]

\addplot[fill=black!85, draw=black!90, nodes near coords style={font=\small\bfseries, text=white}] coordinates {(122, 1) (161, 2)};
\addplot[fill=black!40, draw=black!90, nodes near coords style={font=\small, text=black}] coordinates {(40, 1) (19, 2)};
\addplot[fill=black!10, draw=black!90, nodes near coords style={font=\small, text=black}] coordinates {(38, 1) (20, 2)};

\legend{JaiTTS-v1.0 Wins, Ties, Competitor Wins}
\end{axis}
\end{tikzpicture}
\caption{Head-to-head human judgment results of JaiTTS-v1.0 against commercial flagship models.}
\label{fig:human_judgment_results}
\end{figure*}

\section{Human Judgment Evaluation}

 We select a gender-balanced pool of 30 unique speakers from YouTube (15 female and 15 male) with distinct vocal characteristics. Then, we extract a 10-13 second reference audio sample for each speaker. The models are then tasked with synthesizing speech based on a set of 30 evaluation texts. The text set includes 8 samples containing both code-switching and numbers, 8 samples containing code-switching without numbers, 7 samples containing numbers without code-switching, and 7 pure Thai samples without code-switching or numbers. 

 We recruit 20 native Thai speaker evaluators. For each trial, evaluators listen to the ground-truth reference audio and are presented with two synthesized audios from a randomly selected pair of models without knowing what model the audios are generated from. The presentation order is randomized to prevent position bias, yielding 400 total pairwise comparisons for each model.

 Evaluators are provided with guidelines for determining the winning model based on two primary dimensions: naturalness and intelligibility, and speaker similarity. For naturalness and intelligibility, evaluators are instructed to prioritize synthesized audio that articulates the target text with complete accuracy, including raw inputs containing numbers and code-switching, while also maintaining fluent delivery, realistic human prosody, and a natural rhythmic cadence. For speaker similarity, evaluators are instructed to select the audio that more closely matches the reference prompt in vocal timbre, pitch, and overall acoustic identity.

 We compare JaiTTS-v1.0 against leading commercial systems: the eleven\_v3 model from ElevenLabs\footnote{\url{https://elevenlabs.io/v3}} and the speech-2.8-hd model from MiniMax Speech\footnote{\url{https://www.minimax.io/audio/text-to-speech}}. 
 We select these two models to compare with ours because they are known to be very competitive for many languages including Thai. 

The human judgment results suggest a strong preference for our system. Figure~\ref{fig:human_judgment_results} reports the raw head-to-head voting breakdown for JaiTTS-v1.0 against each commercial competitor. Out of 200 direct comparisons against eleven\_v3, our model wins 161 times, ties 19 times, and loses only 20 times. Against speech-2.8-hd, JaiTTS-v1.0 maintains a lead with 122 wins, 40 ties, and 38 losses. Across all 400 pairwise comparisons, JaiTTS-v1.0 therefore achieves 283 wins, 59 ties, and 58 losses, indicating a strong human judgment preference over current commercial flagships in generating natural Thai speech.

\section{Conclusion}

We introduce JaiTTS-v1.0, a state-of-the-art Thai voice cloning model capable of directly processing raw text containing unnormalized numbers and Thai-English code-switching. Evaluations on our novel benchmark demonstrate that JaiTTS-v1.0 achieves the best CER among evaluated baselines and wins 283 of 400 human judgment comparisons against top commercial systems, all while maintaining a highly efficient Real-Time Factor of 0.11.

\section*{Acknowledgements}
The authors sincerely thank the 20 native Thai-speaking evaluators who dedicate their time to the human judgment tests and provide invaluable feedback for our comparative analysis.

\bibliography{reference}
\end{document}